\documentclass[lettersize,journal]{IEEEtran}
\usepackage[style=ieee,maxcitenames=3]{biblatex}
\usepackage{amsmath,amsfonts}
\usepackage[]{subfigure}
\usepackage{textcomp}
\usepackage{stfloats}
\usepackage{url}
\usepackage{verbatim}
\usepackage{relsize}
\usepackage{graphicx}

\newcommand{\subtitlerelsize}{1} %relative size: integer value
\newcommand{\subtitlelinesep}{0.1em} %line separation: a LaTeX length

\usepackage{relsize}

\usepackage[style=ieee,maxcitenames=3]{biblatex}
\def\x{{\mathbf x}}

\newcommand{\Figref}[1]{Fig.~\ref{#1}}

\def\x{{\mathbf x}}

\def\x{{\mathbf x}}

\title{Helmholtz-Decomposition and Optical Flow:\\[\subtitlelinesep]%
    \smaller[\subtitlerelsize]{}Characterizing slow waves in GCamP recordings}

\begin{document}

\title{Helmholtz-Decomposition and Optical Flow:\\[\subtitlelinesep]%
    \smaller[\subtitlerelsize]{}A new method to characterize GCamP recordings}

\author{Michael Gerstenberger, Dominik Juestel and Silviu Bodea
        % <-this % stops a space
\thanks{D. Juestel leads the Optoacoustic Lab at the Helmholtz-Center Munich, Germany. S. Bodea is Member of the Westmeyer Lab at TU Munich and M. Gerstenberger was Research Associate at Fraunhofer HHI Berlin, Germany. E-Mail the corresponding author at michael.werner.gerstenberger@gmail.com.
}
\thanks{Manuscript submitted 12/2023.}}

%\markboth{IEEE/ACM Transactions on Computational Biology and Bioinformatics,~Vol.~14, No.~8, August~2023}%
\markboth{}
{Shell \MakeLowercase{\textit{et al.}}: A Sample Article Using IEEEtran.cls for IEEE Journals}


\maketitle

\begin{abstract}
During deep sleep and under anaesthesia spontaneous patterns of cortical activation frequently take the form of slow travelling waves. Slow wave sleep is an important cognitive state especially because of its relevance for memory consolidation. However, despite extensive research the exact mechanisms are still ill-understood. Novel methods such as high speed widefield imaging of GCamP activity offer new potentials. Here we show how data recorded from transgenic mice under anesthesia can be processed to analyze sources, sinks and patterns of flow. To make the best possible use of the data novel means of data processing are necessary. Therefore, we (1) give a an brief account on processes that play a role in generating slow waves and demonstrate (2) a novel approach to characterize its patterns in GCamP recordings. While slow waves are highly variable, it shows that some are surprisingly similar. To enable quantitative means of analysis and examine the structure of such prototypical events we propose a novel approach for the characterization of slow waves: The Helmholtz-Decomposition of gradient-based Dense Optical Flow of the pixeldense GCamP contrast (df/f). It allows to detect the sources and sinks of activation and discern them from global patterns of neural flow. Aggregated features can be analyzed with variational autoencoders. The results unravel regularities between slow waves and shows how they relate to the experimental conditions. The approach reveals a complex topology of different features in latent slow wave space and identifies prototypical examples for each stage.
\end{abstract}


\section{Introduction}
\label{sec:intro}

While stimulus dependent activity can be observed during wakefulness, spontaneous patterns of activation dominate in stages of deep sleep \cite{brown2012control}. Similarly, fast neuronal firing is replaced by slow, traveling waves of activation during anaesthesia \cite{steriade1993novel, celotto2020analysis}. Neocortical slow waves can be captured by fluorescence microscopy of GCaMP activity in transgenic mice with a sampling rate of up to 100Hz. 
 
Slow waves represent an important yet highly variable neural phenomenon. Thus methods are required which allow for a systematic measurement of slow wave properties. Previous work that described different types of slow waves relied on manually crafted shape parameters such as the slope of the rising and falling edge of the slow waves in EEG or the spatial extent (local/widespread) of cortical coverage \cite{bernardi2018local}. Some invasive neuroimaging studies have focused on patterns of the spread of activity. For example, the wavefront is measured using delay maps which indicate the temporal offset of an event for every pixel relative to its global onset \cite{celotto2020analysis}.  Townsend and Gong \cite{townsend2018detection} suggest an approach to characterize more diverse temporo-spatial properties of slow waves in GCamP recordings by the kernel-based detection of specific patterns in the vector fields of Dense Optical Flow such as spirals or saddles.\\ We considered related methods for structural MRI \cite{lefevre2009identification}, extend on the idea of an event related slow wave analysis \cite{celotto2020analysis} as well as the one of using Optical Flow \cite{townsend2018detection} and suggest a technique that enables a novel perspective on slow waves as events with source- and sink-regions and specific patterns of directional flow: The combination of Helmholtz-Decomposition and Optical Flow for fluorescence microscopy. It allows to measure properties of events that can incorporate several oscillations even if they have a poor signal-to-noise ratio, occur simultaneously and overlap. The procedure aims to describe slow waves by a dense yet interpretable set of features that includes the shape of the spatially averaged $df/f$, the location of sources and sinks and the direction of flow. To visualize the polymorphism of slow waves and identify prototypical slow waves that are most characteristic for each experimental condition embedding vectors are computed for each event using autoencoders. We test our novel method using a GCamP dataset of mice including six levels of isoflurane. The approach can help in improving our understanding of neural processing in the brain during sleep and under anaesthesia.\\ This paper is structured as follows. In section 2 we briefly explain the mechanisms that generate slow waves during sleep and anaesthesia. Section 3 introduces the approach of using Helmholtz-Decomposition and Optical flow with GCamP data and the results achieved with it are presented in section 4. The last section summarizes and gives a critical discussion on the limitations of the results and the presented approach.

\section{Slow-Waves and Delta Oscillations}

\begin{figure}[ht]
\begin{subfigure}[Oscillatory centers in the brain]{
   \includegraphics[width=1\linewidth]{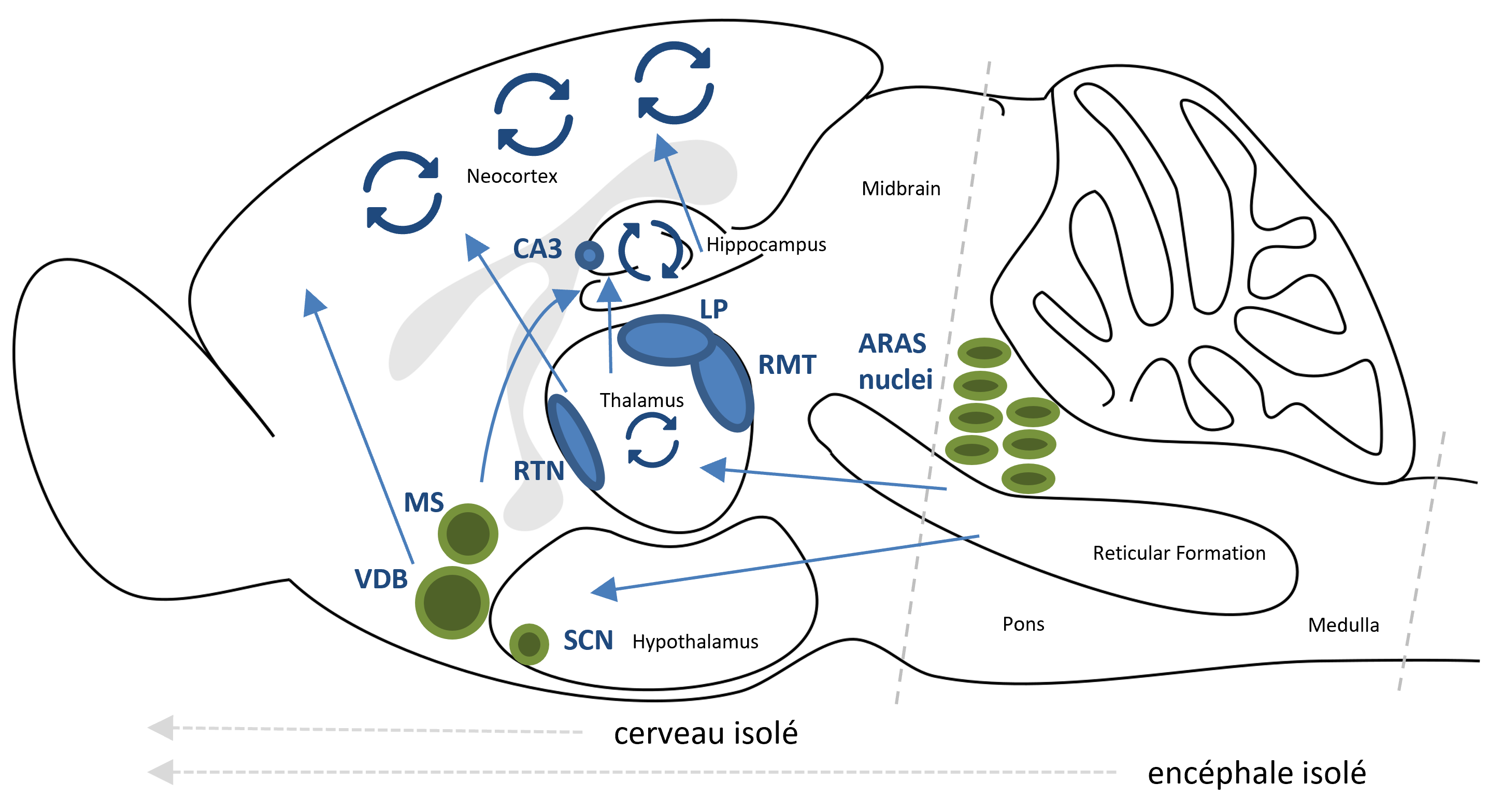}
    }
    \end{subfigure}
\begin{subfigure}[GCamP and ECoG signatures of slow waves]{
   \includegraphics[width=1\linewidth]
      {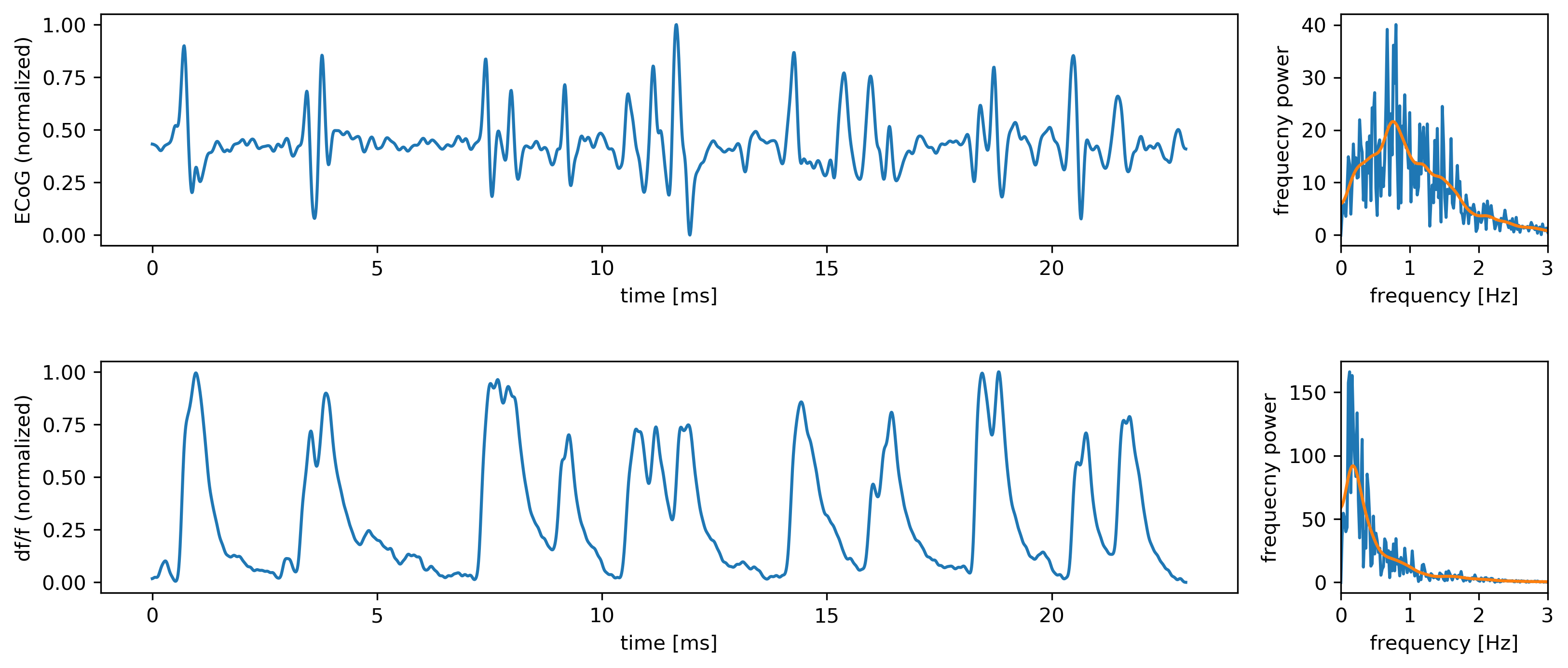}}
\begin{subfigure}[Segmentation of Slow Waves]{
   \includegraphics[width=1\linewidth]
      {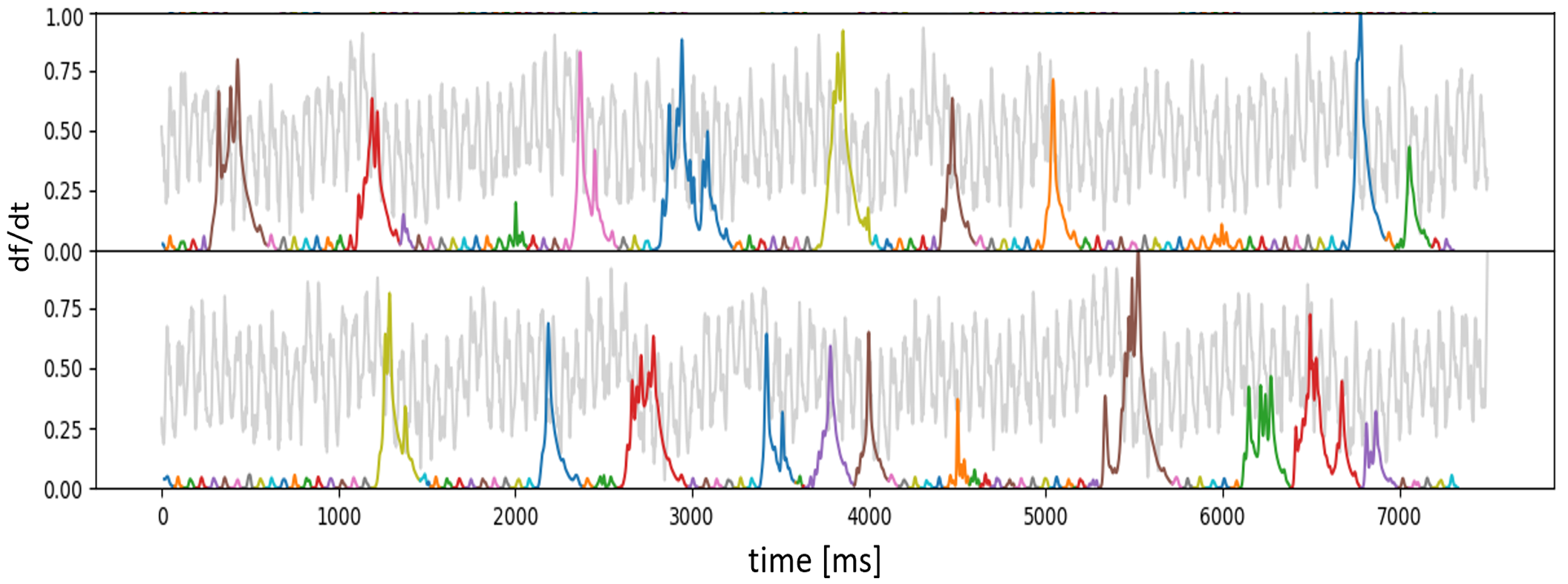}}\end{subfigure}
\end{subfigure}
\caption{Neural oscillators, delta oscillations and slow waves}
\label{overview_slow_waves_and_brain}
\end{figure}

A large body of literature addresses slow oscillations in the brain. Slow waves in the delta range (0.5-4 Hz) can be identified in electroencephalograms (EEG) during deep sleep and under anaesthesia. Neurons that switch between up and down states in the respective frequency exist in thalamus and arguably trigger slow waves. Hence, this kind of slow waves is referred to as thalamocortical slow wave \cite{steriade1984thalamus,brown2012control}. A different type of slow waves that occurs with a frequency below 1Hz has been studied as well\cite{steriade1993novel}: Neocortical slow oscillations. As they do not necessarily occur in regular time intervals they are also referred to as neocortical slow waves here. Note also that the exact frequency depends on the measuring technique. The peak frequency in ECoG tends to be higher as compared to GCamp imaging due to implicit temporal smoothing as a result of the lower sampling rate. This can be shown using publicly available data from \cite{stroh2013making} by computation of the frequency power of ECoG and the aligned GCamP signal. \Figref{overview_slow_waves_and_brain}b shows the spatially averaged df/f and ECoG signals on the left and their Fourier spectrogram on the right. While the peak amplitute is 0.9 Hz in the earlier signal domain it is only 0.2 Hz in the latter.

Both electrophysiological recordings as well as fluorescence microscopy have been used to investigate neocortical slow waves \cite{niethard2018cortical, celotto2020analysis}. Neocortical neurons can oscillate in isolation even in vitro \cite{moghadam2019comparative}. However it was found that slow waves and subcortical signals often occur in an orchestrated way. This holds not only for thalamus. Slow waves arguably also bind together spindles and delta waves \cite{brown2012control}. While named signals are of different origin, one could hence consider them to be the signature of a single, yet distributed process.\\
The question how bistable states arise from the interaction of cells in neural networks has been addressed using spiking network models. They provide an explanation for the occurrence of neocortical slow waves during deep sleep and under anaesthesia. The circumstance that several models explain slow oscillations highlights not only that more then one explanation is possible, but also that different kinds of processes might play a role \cite{nghiem2018two}. While certain characteristics of slow waves differ, bistable states occur both during sleep and anaesthesia in vivo\cite{nghiem2018two} and even spontaneously in vitro \cite{moghadam2019comparative}.

 In contrast to anaesthesia, sleep is a vegetative state. While slow wave sleep is promoted by several mechanisms including the circadian cycle and the release of melatonin that alters neural excitability, sleep spindles of thalamic origin are assumed to be the trigger for a transition to deep sleep \cite{montagna2005fatal}. During slow wave sleep cortex resides in a bistable state with decreased excitability. This decrease in excitation is achieved by a suppression of excitatory input from the ascending reticular activation system (ARAS). As outlined above slow waves may occur spontaneously but can also be triggered by other signals in such states \cite{brown2012control}. The interaction of various neural generators of rhythmic activity is shown by \Figref{overview_slow_waves_and_brain}. Slow wave sleep alternates with REM stages during normal sleep and occurs even in cerveau isolé preparations when excitation from the ARAS is fully absent. A putative function of neocortical slow waves is memory consolidation. The assumed modes of action have been described in the hippocampo-neocortical-dialog model. Empirical results support its core hypothesis that memory replay occurs during sleep and includes both hippocampus and neocortex. Arguably sleep slow waves represent a neural signature of this process\cite{buzsaki1996hippocampo}. 

 The emergence of slow waves under anaesthesia has been studied on a cellular level. Isoflurane has inhibitory effects as it increases K+ channel leakage currents and hence decreases the number of positively charged ions in the cells, making them more negative. Isoflurane anaesthesia manifests in the form of higher outward currents throughout tonic and burst ranges (when K+ channels are active) and a membrane hyperpolarization at rest \cite{ries1999ionic}\footnote{This was demonstrated in thalamocortical cells of the rat using the patch clamp technique. Na+ channels are blocked for control using Tetrodotoxin.}. This means the excitability decreases as a stronger depolarization is required to reach threshold and trigger an action potential. Besides effects on K+ channels a potentiation of glycine receptors is assumed alongside other neurochemical mechanisms that alter the excitability of neurons \cite{pubchem2020iso}. While isoflurane is known to reduce the CL- influx upon administration of GABA \cite{jenkins1999effects} and hence decreases its inhibitory effect, electrophysiological studies in vivo indicate that the net-effect of isoflurane is inhibitory for all relevant dosages. In this respect, isoflurane contrasts with other anaesthetics including halothane and ketamine that show concentration dependence \cite{moghadam2019comparative}. Isoflurane hyperpolarizes cells which inhibits neural signal transduction by a reduced gap junctional conductance. 

While named effects explain the inhibition of neural firing under anaesthesia, decisive effects arise on the population level. Eger (1981) systematically studied the EEG patterns in the awake state and during anaesthesia with isoflurane for five different dosages at up to 2.9\% in humans. For very light anaesthesia (iso = .56\%) low voltage fast activity can be observed. At a light surgical level (iso = .96\% and 1.78\%) slow oscillations are present that change from more regular to irregular. Alternating patterns with high amplitude oscillations can be observed at a moderate surgical level (iso = 2.2\%). For deep anaesthesia only occasional low voltage activity shows (iso = 2.9\%). Isoflurane administration leads to a gradual shift from a stable awake state to bistable states and finally deep anaesthesia where quiescence dominates.\\
In summary it shows that neocortical slow waves emerge spontaneously in cortex while they can be also be triggered by signals from subcortical structures. Methods for the characterization of slow waves with widefield fluorescence must help in detecting the source regions of slow waves and describe how the patterns spread and where they target. The circumstance that a complex dynamic occurs on a population level that manifests in characteristic sequences of oscillations during the up-state of a slow wave reflect how cortical pathways act as a generator of rhythmical activity.

\section{Method}
We present a new approach for the analysis of slow waves that includes three steps: (1) Detection
of events, (2) feature extraction by Optical Flow and Helmholtz-Decomposition
as well as (3) the retrieval of embedding vectors by aggregation and the use of Autoencoders for the visualization of the slow wave distribution and the detection of prototypical events.
%The suggested procedure includes four steps. First the slow waves are detected, second vector fields are estimated using Dense Optical Flow, third features are measured using Helmholtz-Decomposition and fourth latent embeddings are computed
\begin{figure}
\begin{subfigure}[Sources, Sinks and Flow]{
   \includegraphics[width=1\linewidth]{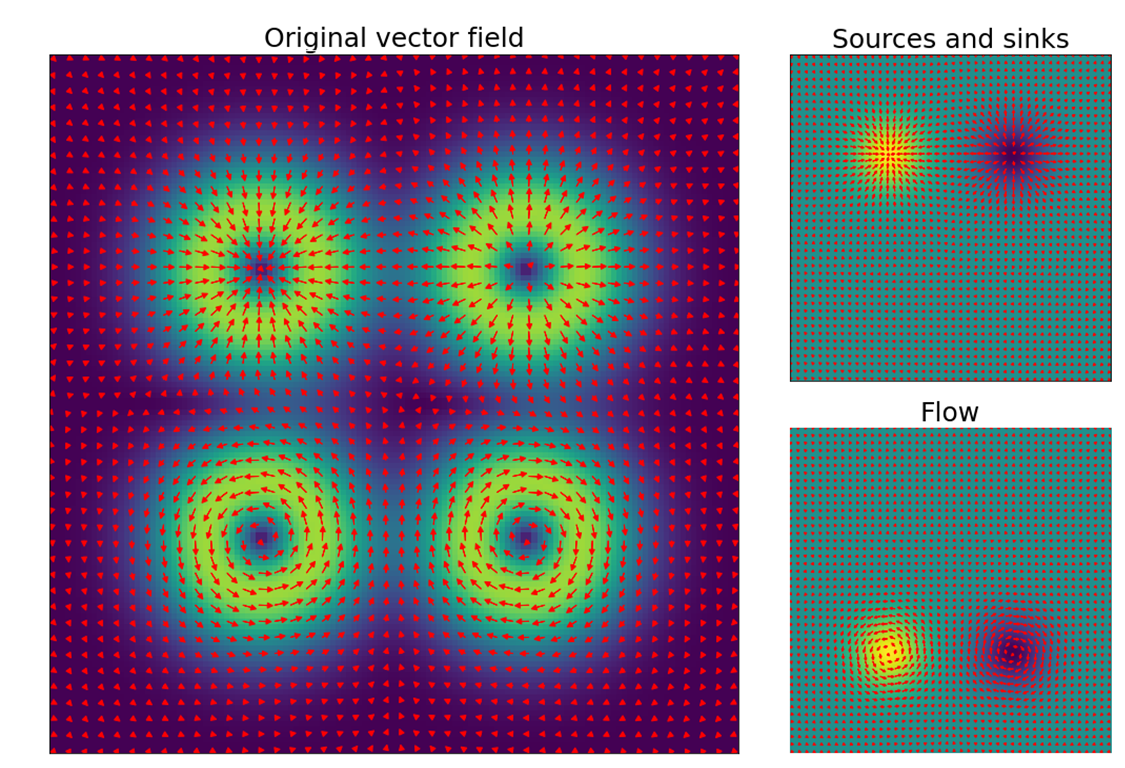}
}\end{subfigure}
\begin{subfigure}[Helmholtz-Decomposition for simulated Optical Flow]{
   \includegraphics[width=1\linewidth]
{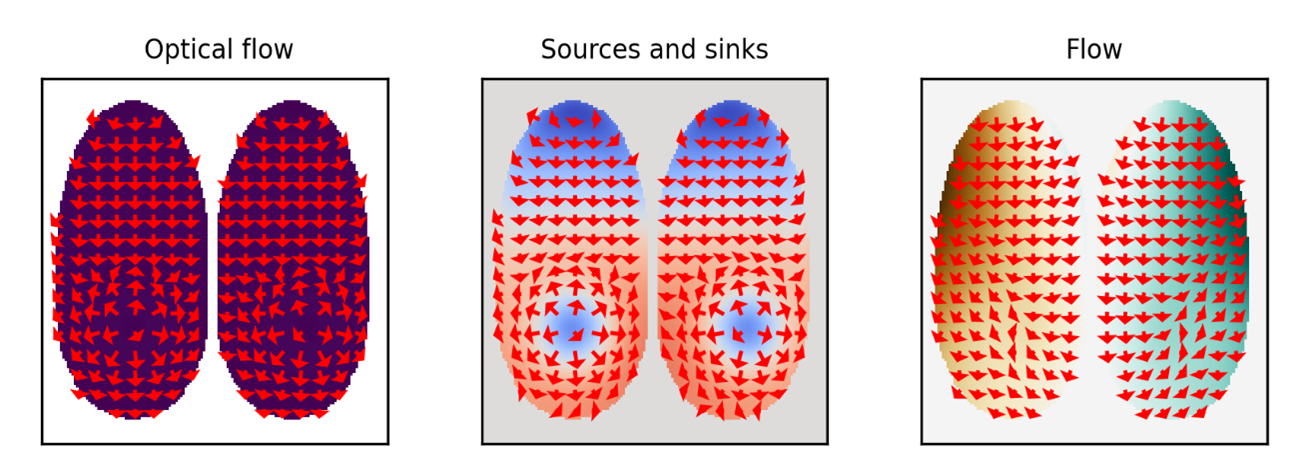}
   }
\end{subfigure}
\caption{Helmholtz-Decomposition of Optical Flow}
\label{helmholtz_decomposition}
\end{figure}

To detect slow waves we compute the $dF/F$ signal for each recording as $dF/F =( F_t - F_0)/F_0$. We refine the signal by band-stop filtering to remove effects of the heartbeat ($10$-$20\text{Hz}$) and take the arithmetic mean over the spatial domain. After detrending, the onset/offset of the events can be determined by thresholding the resulting 1D time series(\Figref{overview_slow_waves_and_brain}c). We continue with the computation of the pixeldense $dF_w/F_w$ signal per slow wave i.e. the percentage change of each frame $f_{wt}$ as compared to baseline activity $F_{w0}0$ of the event.
\begin{equation}
dF_w/F_w =( F_{wt} - F_{w0})/F_{w0}    
\end{equation}
 Vector fields of pixel displacements can be computed from fluorescence recordings using gradient based Dense Optical Flow\cite{townsend2018detection}. To this end we rely on the method suggested by Horn and Schunck\cite{horn1981determining}. It estimates the Optical Flow using both the image gradient and the temporal derivative between the pixel values of two subsequent frames. Optical Flow is computed by iterative minimization of an objective function $e$. The first part $e_f$ of this function measures the error in the optical flow constraint of the vector fields.\\
\begin{equation}
    e_f=\iint (I_xu + I_yv + I_t)^2{{\rm d}x{\rm d}y}
\label{eqn:horn_schunck_1}
\end{equation}

Here $I_x$ and $I_y$ are the spatial derivatives of the $df/f$ signal, $u$ and $v$ denote the vector components of the flow field to be computed and $I_t$ is the temporal derivative. Recall that optical flow can be measured uniquely in 1D but the solution in 2D is ubiquitous (aperture problem). Hence an additional assumption is made and a global smoothness constraint introduced by the Horn-Schunck method.
\begin{equation}
e_s=\iint (\lVert\nabla u\rVert^2+\lVert\nabla v\rVert^2){{\rm d}x{\rm d}y} 
\label{eqn:horn_schunck_2}
\end{equation}

The energy functional is the sum of both terms where the global smoothness is weighted by the parameter $\alpha^2$.

\begin{equation}
E = e_f + \alpha^2 e_s
\label{eqn:horn_schunck_3}
\end{equation}

An iterative scheme can be derived to compute the vector fields\cite{horn1981determining}. We suggest computing Optical Flow separately for each hemisphere and then masking the background vectors. The resulting vector fields reflect the apparent motion and approximates the displacement of pixels with the same intensity (GCaMP fluorescence) between frames of the $df/f$ contrast. Hence, it implicitly captures the direction and speed at which neocortical slow waves spread.
%$E=\iint \left[(I_xu + I_yv + I_t)^2 + \alpha^2(\lVert\nabla u\rVert^2+\lVert\nabla v\rVert^2)\right]{{\rm d}x{\rm d}y}$
%https://slideplayer.com/slide/16093197/88/images/15/Computing+Optical+Flow%3A+Horn-Schunck+method.jpg

Patterns in these vector fields can be used to characterize the way neural signals travel in cortex during slow wave sleep and anaesthesia. Townsend and Gong (2018) suggest a kernel-based approach to detect several local and global features. A more general account that allows to study fundamental properties of vector fields is Helmholtz-Decomposition. The fundamental theorem of vector calculus states that every vector field can be expressed as the sum of a curl free and a divergence free vector field if it is sufficiently smooth and decays rapidly.

\begin{equation}
v=\mathrm{\nabla}\varphi_{curl=0}+J\mathrm{\nabla}\varphi_{div=0}
\label{eqn:helmholtz_decomposition}
\end{equation}

 Helmholtz-Decomposition allows to retrieve the curl free and the divergence free component for the given vector field. The divergence free vector field is also referred to as flow (\Figref{helmholtz_decomposition}a). The vector fields are expressed via the gradient $\nabla\varphi$. J relates to a matrix that rotates the gradient of the scalar potential of the divergence free component by 90 degrees (Eq. \ref{eqn:helmholtz_decomposition}). The flow component of displacement vectors indicates the global dynamics of neural activity and shows the overall direction of the slow waves. 
   \begin{equation}
  \begin{array}{l}
  O_{\text{sources}}=
  \begin{cases}
      \mathrm{\nabla}\varphi_{curl=0},& \text{if } \mathrm{\nabla}\varphi_{curl=0} > 0\\
      0,              & \text{otherwise}
  \end{cases}\\
  I_{\text{sinks}}=
  \begin{cases}
      \mathrm{\nabla}\varphi_{curl=0},& \text{if } \mathrm{\nabla}\varphi_{curl=0} < 0\\
      0,              & \text{otherwise}
  \end{cases}
  \end{array}
  \label{eqn:sources_versus_sinks}
\end{equation}
 The curl free vector field can be separated into the distribution of sources $O$ and sinks $I$. Sources are especially important because they indicate the origin of focal activity. \Figref{helmholtz_decomposition}b illustrates the effects of the Helmholtz-Decomposition of Optical Flow for a simulated signal with increasing global $df/f$ originating from the top and a focal spot of raising intensity(\Figref{helmholtz_decomposition}b; left). The constructed global downwards trend is captured by the flow component (\Figref{helmholtz_decomposition}b; right) while the location of the slow wave origin manifest as sources visible as blue areas in the scalar potential (\Figref{helmholtz_decomposition}b; left).\\
 The suggested approach allows to assess properties of slow waves quantitatively. To this end we compute several derived measures: The vertical fraction of flow $\textit{Vertical flow} = u_{\text{flow}}/(u_{\text{flow}}+v_{\text{flow}})$, the share of bottom-up flow of vertical flow $\textit{Bottom up} = u_{\text{flow}>0}/u_{\text{flow}}$, the fraction of medial to lateral flow ($v_{\text{flow}>0}/v_{\text{flow}}$) for the left and $v_{\text{flow}<0}/v_{\text{flow}}$ for the right hemisphere),
the total upwards- ($\sum{u_{\text{flow}>0}}$) and downwards flow ($\sum{u_{\text{flow}>0}}$) as well as the leftwards- ($\sum{v_{\text{flow}<0}}$) and rightwards flow ($\sum{v_{\text{flow}<0}}$) of the left hemisphere, the peak amplitude of the GCamP signal $\max df/f$ its duration as well as the temporal integral over the sources $\int O {\rm d}t/n$ and sinks $\int I {\rm d}t/n$ that shows the average distributions per event. These features capture important characteristics of slow waves that can be displayed and interpreted on the basis of individual events.
\\
We test using variational autoencoders (VAEs) to retrieve a low dimensional latent embedding and investigate distributions of different types of slow waves exploratively. The position on the embedding manifold reflects what the events look like and thus shows characteristic features of the events. This allows for an event related analysis of slow waves and to investigate differences between conditions.\\
\begin{equation}
L = \text{KL}_{loss}(z_{mean},z_{logvar}) + \sum p_i \text{MSE}(\text{d}_i, \text{VAE}(d_i))
\label{eqn:custom_loss}
\end{equation}

 \begin{figure}
   \includegraphics[width=1\linewidth]{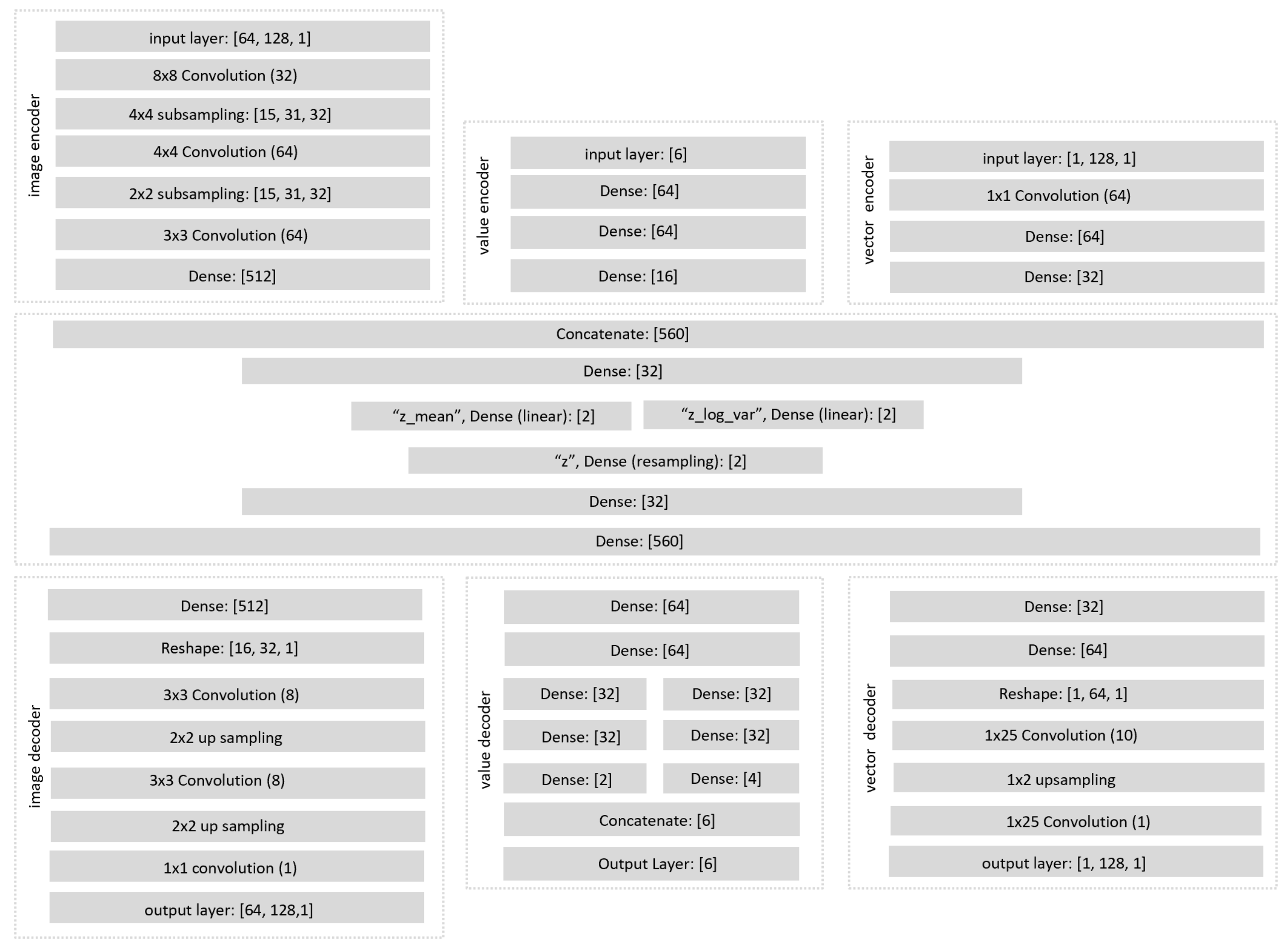}
\caption{A mixed input Variational Autoencoder (VAE)}
\label{autoencoder_2}
\end{figure}

  The first VAE embeds the GCamP signal $\int df/f {\rm d}t/n$ using six fully connected layers with ReLU activation. It reveals the topology of slow waves according to the shape of the spatially averaged GCamP signal (see \Figref{slow_wave_shape_space}). A second model embeds a larger number of slow wave features. They include the spatial distribution of the sources $\int O {\rm d}t/n$ and sinks $\int I {\rm d}t/n$, the GCamP time series that was averaged spatially, the duration and amplitude of the event as well as the direction of flow. The latter is aggregated in four values that represent the sum of the upwards-, downwards-, leftwards and rightwards vector components of the flow field that was computed using the Helmholtz Decomposition of the Optical Flow of the df/f signal (see \Figref{latent_slow_wave_space}). Our third model auto-encodes the GCamP signal $\int df/f {\rm d}t/n$ as well as the upwards and downwards flow. A custom loss function was used to ensure that all of the encoded features are well reflected in the reconstructions. The respective function is given by Equation \ref{eqn:custom_loss} where MSE relates to the mean squared error and $d$ is a set of the inputs to the respective model.  \Figref{autoencoder_2} depicts the mixed input model. The parameter $p_i$ of the individual terms in the loss functions were adjusted manually while changes to the network structure were made until all the predictions correlated strongly with the data.\\  Prototypical examples are detected with a class specific GMM that approximates the conditional likelyhood of the embedding vectors $z$ of each condition $c=1\ldots5$ by an independent set of Gaussian mixture components ($k=3$).
  \begin{align*}
    p_c = \sum_{k=1}^Nw_k{}c \mathcal{N}(\mu_k{}c,\Sigma_k{}c), \text{where}\\
    Z | K \sim \mathcal{N}(\mu_K{}c, \Sigma_K{}c)
\end{align*}
Each mixture component corresponds to a prototype and its mean $\mu$ represents the prototype vector. The prototypical samples are the ones with the embedding vector that has the highest likelyhood for the respective gaussian kernel. 

\section{Results}
\begin{figure}[ht]
   \includegraphics[width=1\linewidth]{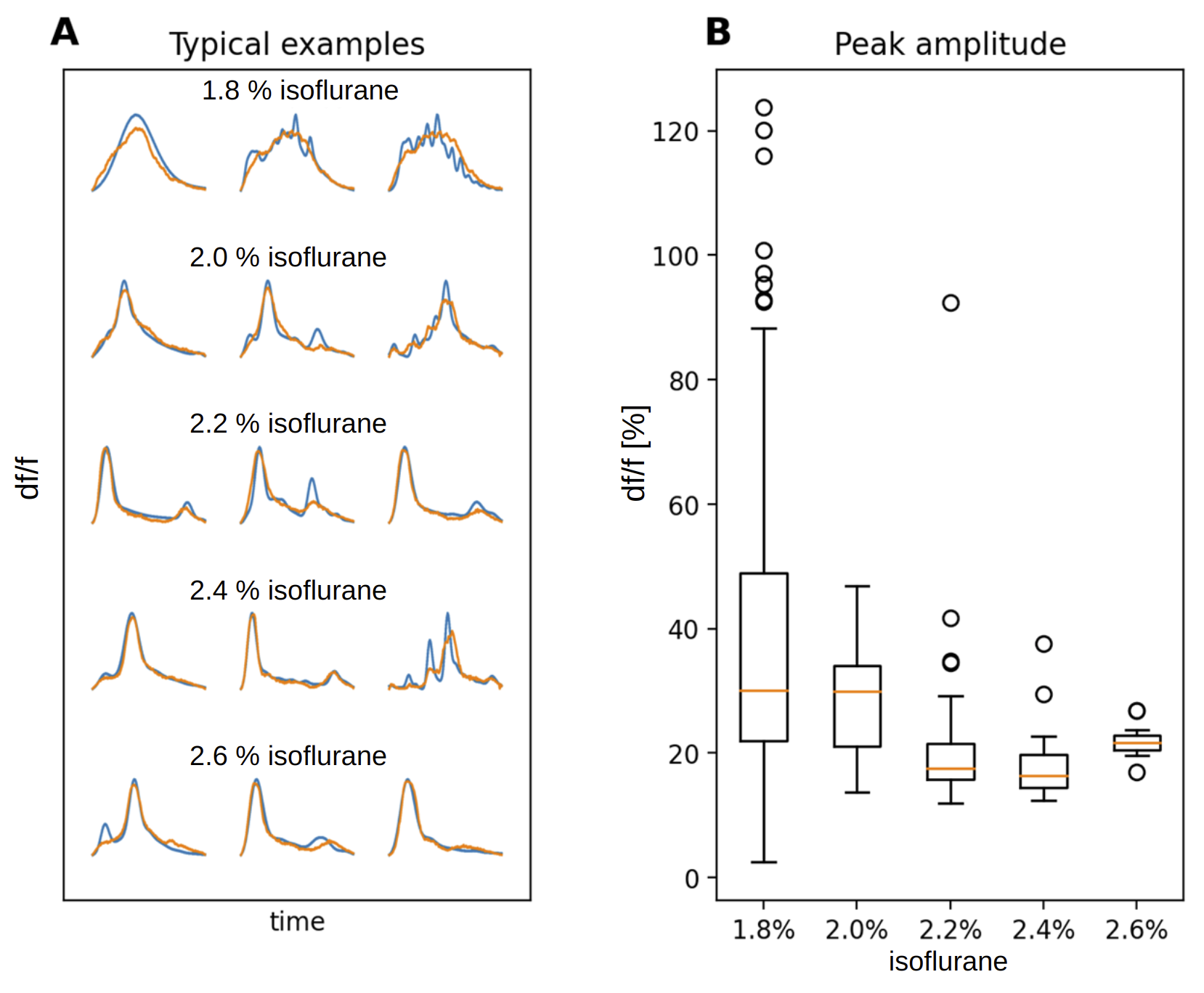}
    \caption{Prototypical Slow Waves}
    \label{typical_examples_and_peak_amplitudes}
\end{figure}

\begin{figure}[ht]
   \includegraphics[width=1\linewidth]{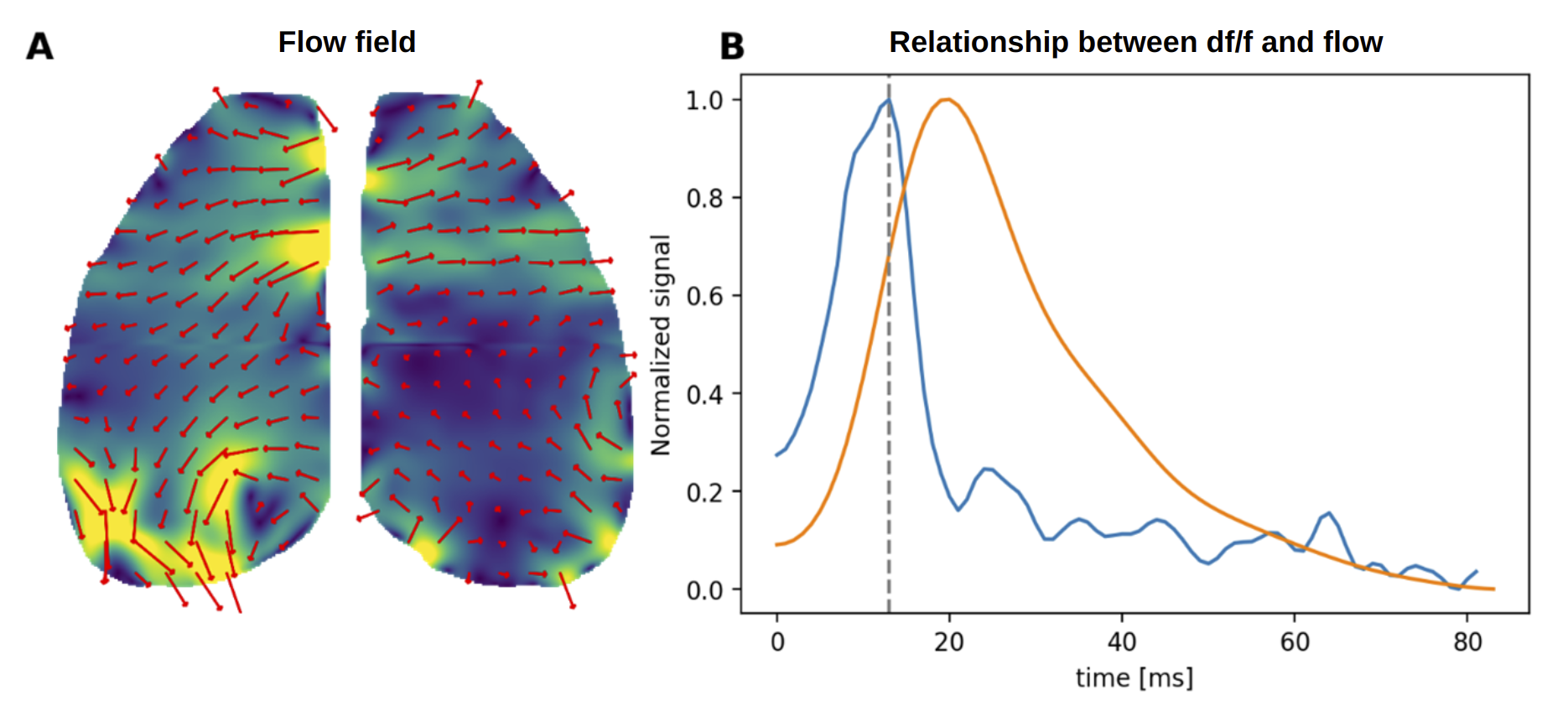}
    \caption{Vector Field of Flow}
    \label{vector_field_flow_component}
\end{figure}

We apply the proposed method on a dataset from an experiment with five conditions. Each recording has a length of 30 seconds and was measured at a different level of isofluarane ranging from 1.8\% to 2.6\%. Several different types of neocortical slow waves can be discerned intuitively based on the shape of the temporal mean of the respective df/f signal. Events with peak-amplitude below 5\% in the df/f signal and a high correlation with the hemodynamic signal ($r>.3$) were excluded from the analysis to prohibit potential artifacts from affecting the analysis that result from variations in the oxygen levels due to breathing.\\ \Figref{typical_examples_and_peak_amplitudes}A shows the prototypical examples of the detected slow waves in the embeddings of VAE 1 for the different conditions in blue and the reconstructions in orange. While longer periods of continuous activity with multiple peaks can be observed at 1.8\% isoflurane, increasing levels of anesthetics correlate with events that have few peaks or even only a single maximum. As a tendency the average amplitude decreases with higher levels of isoflurane. The same holds for the standard deviation of the amplitudes (\Figref{typical_examples_and_peak_amplitudes}B).\\
As described in the method section the results of the Helmholtz-Decomposition of Dense Optical Flow can be well visualized. 
\Figref{vector_field_flow_component} illustrates the dynamics of flow during an event with a single peak and reveals its relationship to the $df/f$ signal. Visual inspection of the $df/f$ videos gives the impression of an event where activity spreads from frontal-medial areas of the left hemisphere towards lateral and parietal portions of cortex. This pattern is well captured by the flow component. \Figref{vector_field_flow_component}A shows the vector field for the measurement with the maximal global field strength ($d_t = 18ms$). Arrows indicate the direction and strength of flow, the latter of which is also indicated in the background. \Figref{vector_field_flow_component}B shows the relationship with the $df/f$ signal. The flow is strongest in the beginning of the event during the rising phase of $df/f$ which continues to increase thereafter. While activity spreads in early stages of the slow wave a uniform change of fluorescence that involves all cortex typically occurs later.
\\

Areas with a focal increase in GCamP activity have high values in the scalar potential of the sources (\Figref{helmholtz_decomposition}b). Sources indicate the origin of neural activity as indicated by the Optical Flow of the $df/f$ signal. For many single peak events there are distinct source areas, directional flow that originates from them and targets at sinks where activity prevails longest. 
\begin{figure}[ht]
   \includegraphics[width=1\linewidth]{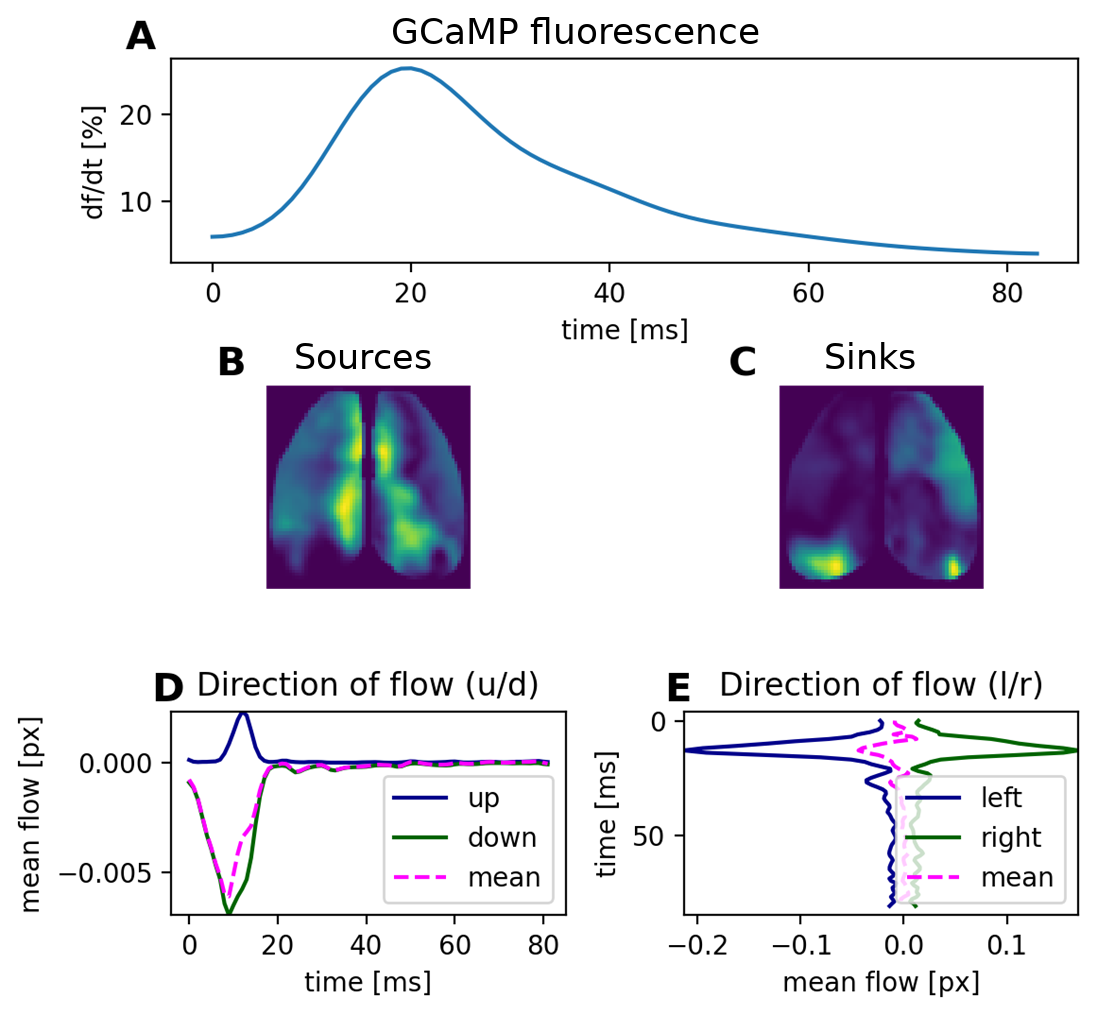}
    \caption{Sources, sinks and flow}
    \label{sources_sinks_and_flow}
\end{figure}
For one type of slow waves medial portions of cortex typically act as the sources while occipital areas are the sinks (\Figref{sources_sinks_and_flow}B and C). These events are characterized by pronounced media-to-lateral flow (\Figref{sources_sinks_and_flow}E) and a tendency towards downwards flow (\Figref{sources_sinks_and_flow}D). Visual inspection of the $df/f$ videos confirms this dynamic of cortical activity that originates in frontal areas, spreads quickly downwards towards medial areas and then laterally. It shows empirically that activity in areas that act as a sink typically lasts longest.

\begin{figure}[ht]
   \includegraphics[width=1\linewidth]{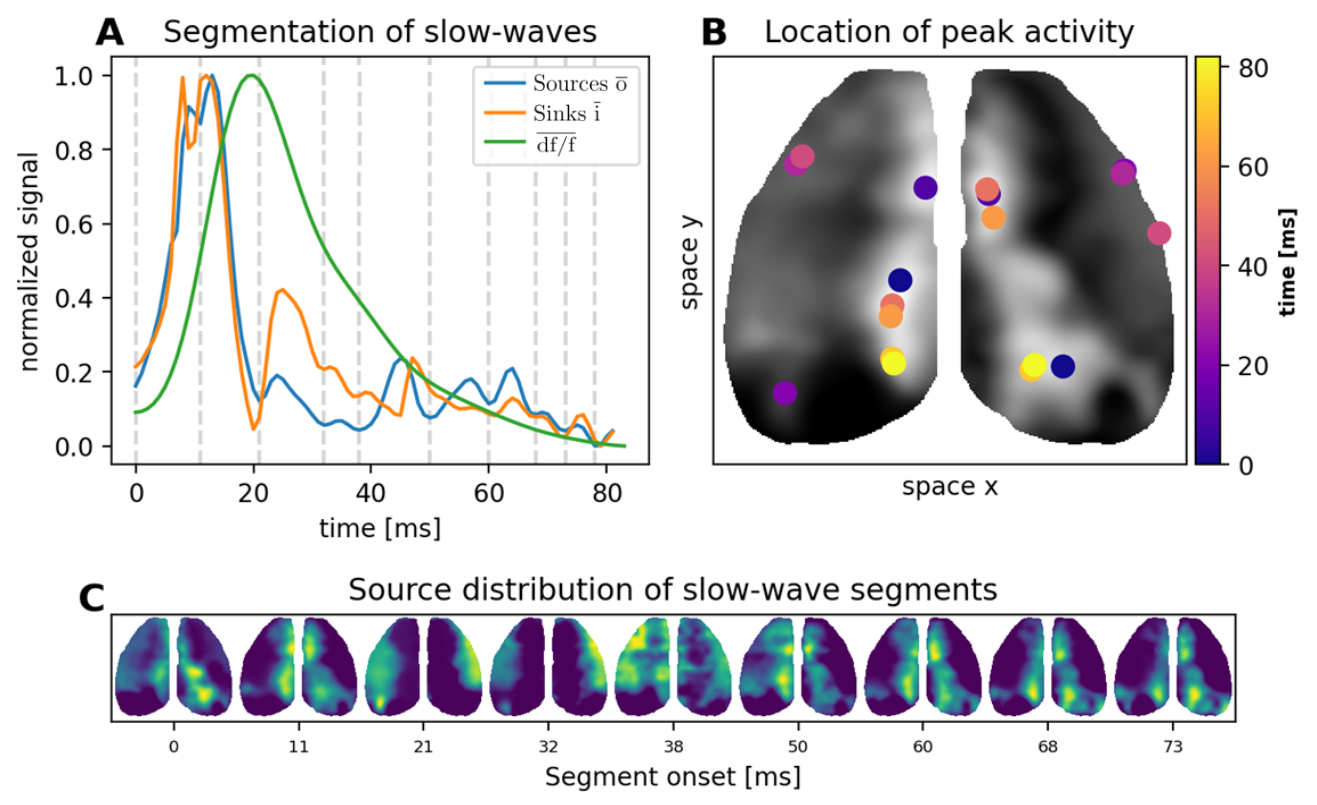}
    \caption{Subsegments of a slow wave}
    \label{fig:my_label}
\end{figure}

\begin{figure}[ht]
   \includegraphics[width=1\linewidth]{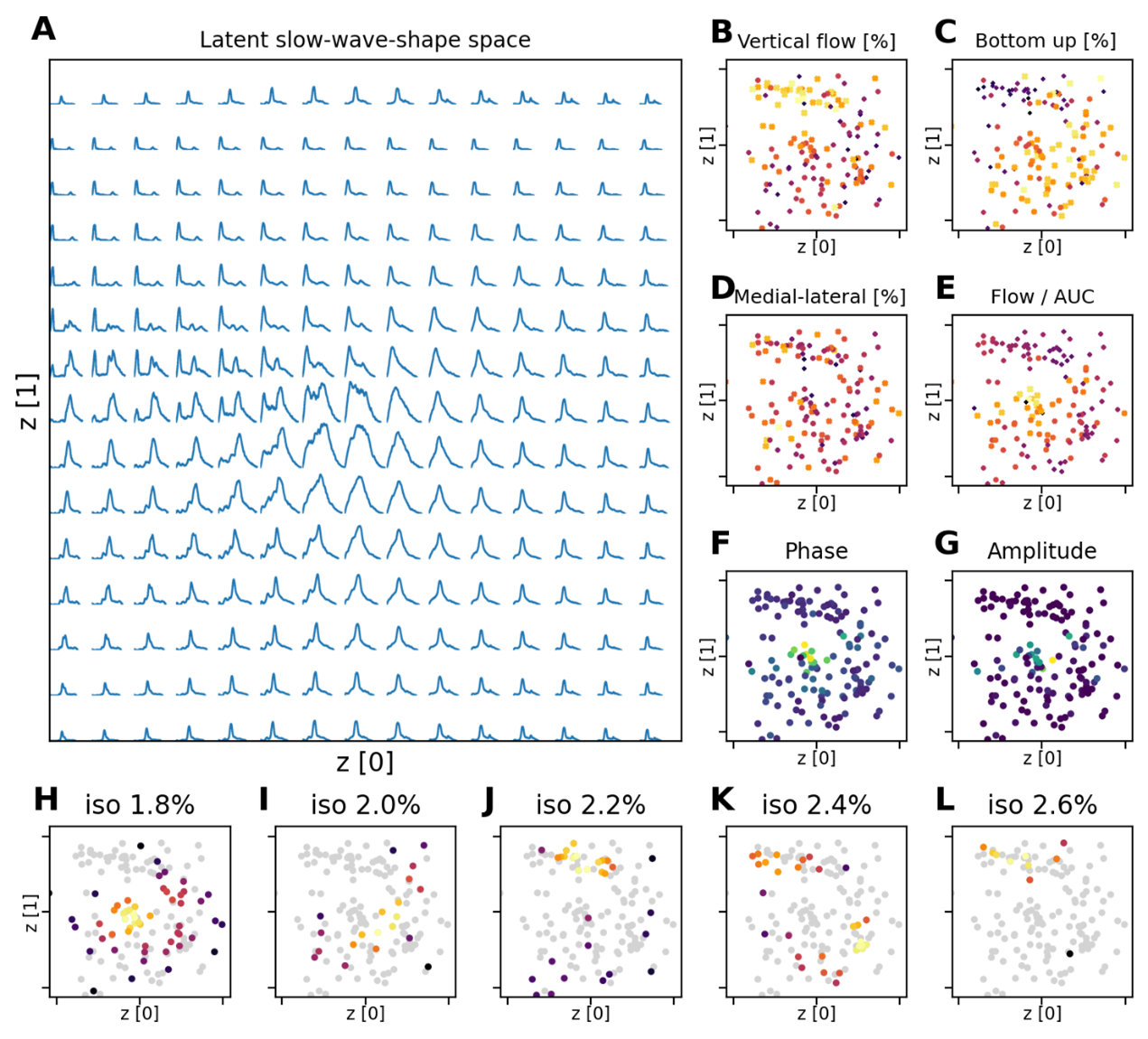}
   \caption{Slow wave shapes in feature space (VAE 1)}
   \label{slow_wave_shape_space} 
\end{figure}

We encode different sets of features using variational autoencoders to inspect the distribution of samples with different isoflurane levels in latent space. Events with peak-amplitude below 5\% signal change or high correlation with breathing (r$>$.3) were excluded from the analysis. The first model targets at a latent space defined by the slow wave shape. \Figref{slow_wave_shape_space} illustrates the results. Panel A shows a reconstruction manifold for latent-layer activations chosen from an evenly spaced grid. Larger waves are mapped to the center while smaller ones lie in the periphery: The phase and amplitude is higher for the respective samples (Panels F-G). The samples in the center region also have stronger flow as measured by the area under the $df/f$ curve. Samples from the different experimental conditions are unevenly distributed. Panels H-L show the distribution of samples with different isoflurane levels. The density of the distribution for overlapping points is color-coded. All samples that do not belong to the respective condition are plotted in grey. Most samples with an isoflurane level of 1.8\% lie within the center and represent events with multiple peaks and longer upstate periods. In contrast the examples at 2\% lie further to the lower right and represent slow waves with one peak and a steep rising and falling edge. For higher levels of isoflurane a cluster exists in the upper right. This cluster arguably corresponds to waves with two peaks the smaller one of which corresponds to an oscillation that is triggered by breathing. A relationship between breathing and neural oscillations has been identified before \cite{tort2018parallel}. These samples have strong vertical flow (as compared to horizontal flow) and little bottom-up flow (as compared to top-down flow; Panels B-D). Other samples of the very deep anaesthesia conditions (iso$>$2\%) are mapped to to the lower-right. This indicates a general shift from longer lasting up states with several oscillations of varying  to less dynamical states where smaller waves typically have only one single peak.

\begin{figure}[ht]
   \includegraphics[width=1\linewidth]{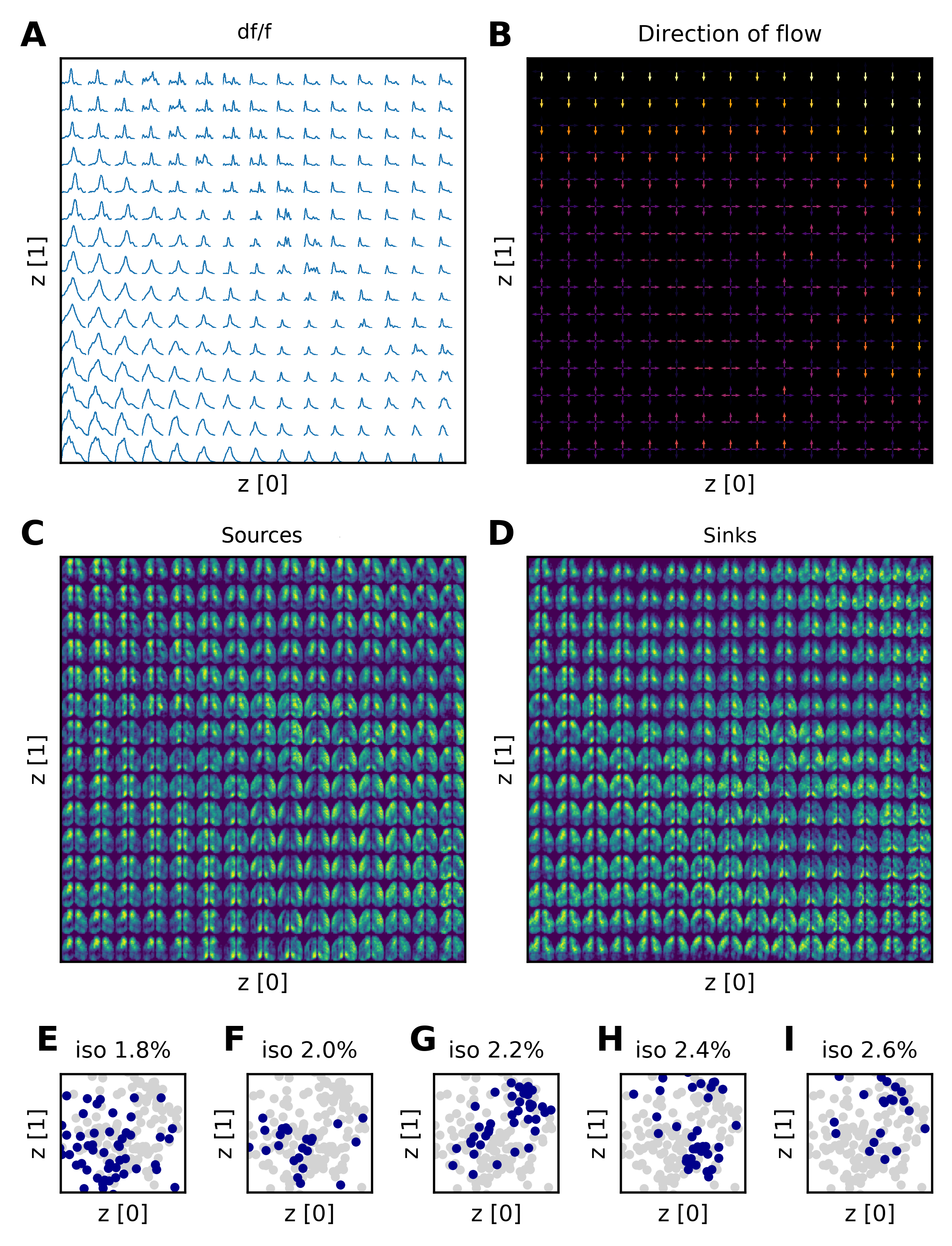}
   \caption{Slow waves in a combined feature space (VAE 2)}
   \label{latent_slow_wave_space} 
\end{figure}

\Figref{latent_slow_wave_space} shows the results of for the mixed input model. Panel A shows the reconstructions of the df/f signal. Duration and amplitude are square root scaled. Panel B indicates the direction of flow. Note that upwards flow dominates in the lower center, flow occurs in all directions rather evenly in the middle left portion while downwards flow dominates for events in the upper right area. Panel C and D show the reconstructions of the temporal average of the sources and sinks. Panels E to I show the distribution of samples of different experimental conditions in latent slow wave space. \\
The model encoders the shape of the df/f signal alongside the duration and amplitude, the ratio between flow in different directions as well as the distibution of sources and the distribution of sinks. Hence latent slow wave space allows for a more complete distinction of neocortical slow waves. Although only a two dimensional latent space was used an acceptable reconstruction was achieved for all included features. Here large amplitude waves with multiple peaks that are typical for low isoflurane levels (iso=1.8\%) are mapped to the lower left corner. Within this area one may discern events with frontal sources (and ubiquitous flow) and those with sources along the medial band (with flow rather bottom-up or medial-to lateral). Large single peak waves that are characteristic for 2\% isoflurane are mapped to the central portions of the lower left quadrant - an area with mostly medial to lateral or upwards flow. Small waves that include a second peak lie in the upper right area. Sources in frontal areas are present for small waves that exist mainly during deep anaesthesia (iso$>$2.2\%). This type of small waves is characterized by downwards flow.
The manifold of sources and sinks provides further information regarding the nature of larger waves. Neocortical slow waves that are mapped to a region to the lower right have sources in areas of the brain that potentially relate to the barrel fields in fronto-peripheral parts of cortex. Especially for the latter kind of events sinks reside in areas different from the sources and upwards flow dominates.
\\
\begin{figure}[ht]
   \includegraphics[width=1\linewidth]
      {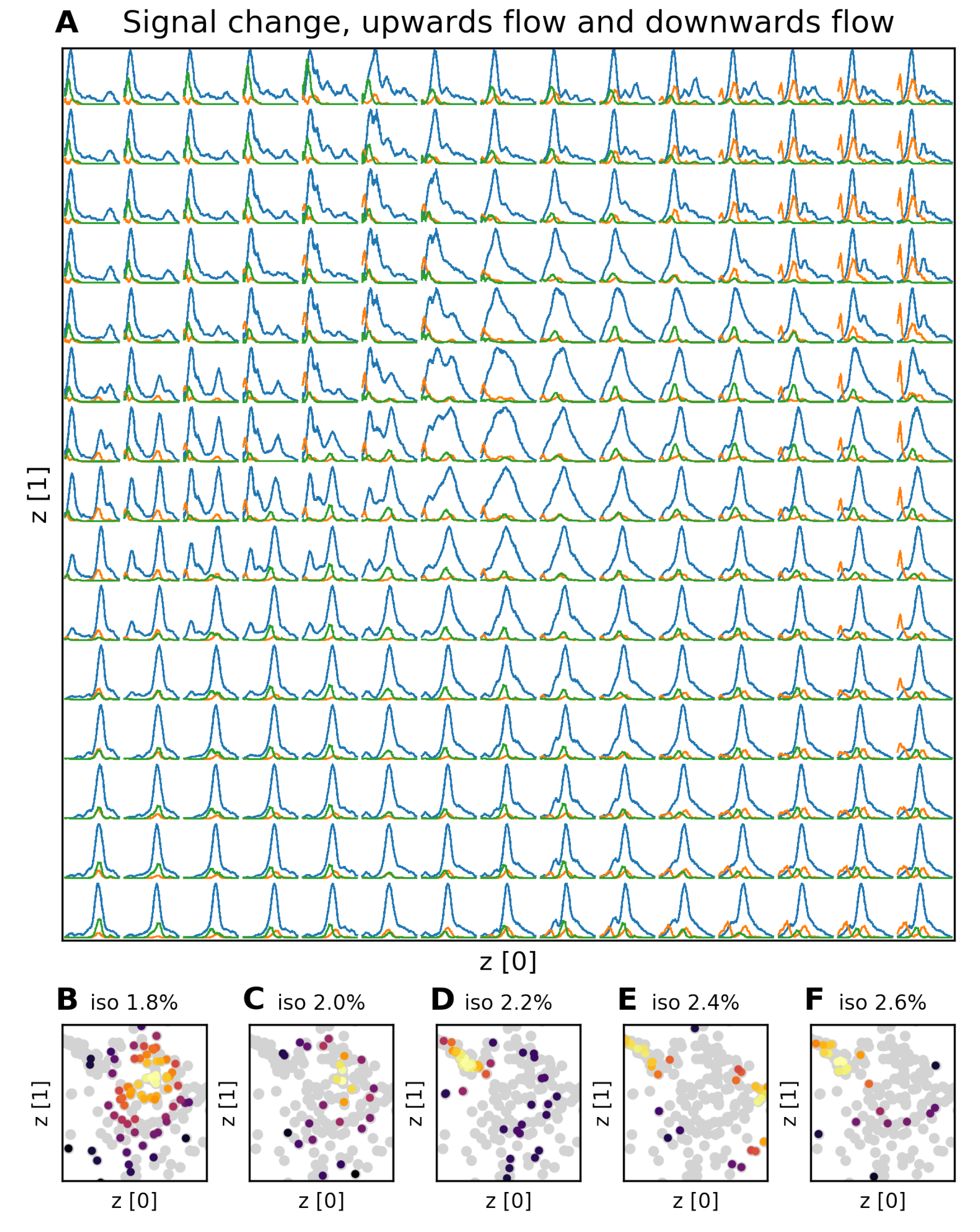}
   \caption{Latent distribution of Flow (VAE 3)}
   \label{fig:Ng3}
\end{figure}
The third model auto-encodes the $df/f$ signal alongside the upwards and downwards flow and can be used to examine regularities in the direction of flow during different stages of slow waves. It shows that in most cases there is a point in time when most upwards or downwards flow occurs. This holds typically for the first rising edge of the events. At 1.8\% isoflurane many events begin with downwards flow (orange peak; central reconstructions) while pronounced peak in upwards flow (green peak; central portion of upper right) shows in areas of latent space mainly occupied by examples of 2\% isoflurane. This tendency for bottom up flow also holds for events at higher levels of isoflurane (upper left).\\
Our results reveal a complex topology of events. It reveals the potentials of the proposed method. The analysis with VAEs is especially well suited to understand the structure of the data exploratively. We found several types of slow waves for the given dataset and provide the prerequisites for a systematic, hypothesis-driven analysis of slow waves with fluorescence microscopy. The combination of Helmholtz-Decomposition with the Optical Flow of the pixel-dense GCamP contrast (df/f) represents a viable tool for further research.

\section{Summary and Discussion}
We demonstrate a new method to characterize neocortical slow waves using fluorescence microscopy. It relies on the Helmholtz-Decomposition of Dense Optical Flow and captures the global dynamics of spread. Moreover it allows to capture the sources and sinks of of the events. They can be interpreted as the sites of origin of neocortical slow waves and the areas activity converges at and prevails longest. Thus it captures how different brain regions interact during bistable states of anaesthesia and arguably also during deep sleep. Our results appear promising and show the potentials of the approach. Convolutional VAEs have proven useful to retrieve a low dimensional latent representation that can be used to visualize feature spaces that describe the variability of slow waves. Prototypical events can be identified by a categorical GMM. We understand slow waves as periods of an extended up-state that may incorporate multiple oscillations in different regions of the brain. Therefore they reflect how activity travels along functional cortical pathways which might play an important role as a generator of rhythmical activity in neocortex. The Helmholtz-Decomposition of the Optical Flow of the pixeldense GCamP contrast is well suited to detect important features of neural activity and characterize slow waves.\\
The proposed method can be used to characterize and distinguish slow waves that occur at different stages of anaesthesia. We tested with recordings acquired from a single transgenic mouse. The analysed data with a high sampling rate of (100Hz) was measured for time frame of 30s per condition. Although the number of events is thus limited the results are congruent with the changes in cortical dynamicd reported previously. We found that different types of slow waves form clusters with respect to different features. At low isoflurane concentrations events typically have multiple peaks and a complex dynamic that incorporates periods of downwards and subsequent upwards flow. With increasing levels of isoflurane the amplitude decreases and shorter events with a single maximum prevail. One type of events appears to be triggered by the somatosensory state of the animal. This type of low amplitude events that are phase locked to breathing occur at deep anesthesia. Several regions act as sources of neural activity. Frontal areas are most common for lower levels of isoflurane while, medial, occipital or lateral clusters become more frequent at higher concentrations that cause deep anaesthesia. \\
Future research should focus on means of a fine grained categorization of different types of slow waves that was not attempted here. Considering the high temporal resolution of modern GCamP recording setups and the interactions of cortical and subcortical sites during bistable neurological states the approach offers potentials not only for studiyng slow wave anaesthesia but potentially also the interaction of different brain regions and rhythmical activity during sleep.\\

\printbibliography[heading=bibintoc]

\end{document}